\documentclass{article}

\PassOptionsToPackage{number, compress}{natbib}

\usepackage[nonatbib]{neurips_2023}

\usepackage[utf8]{inputenc} 
\usepackage[T1]{fontenc}    
\usepackage{hyperref}       
\usepackage{url}            
\usepackage{booktabs}       
\usepackage{amsfonts}       
\usepackage{nicefrac}       
\usepackage{microtype}      
\usepackage{xcolor}         

\usepackage{multirow}
\usepackage{amssymb}
\usepackage{pifont}
\usepackage{amsmath}
\usepackage{aligned-overset}
\usepackage{xspace}

\usepackage[capitalize]{cleveref}
\crefname{section}{Sec.}{Secs.}
\Crefname{section}{Section}{Sections}
\crefname{table}{Tab.}{Tabs.}
\Crefname{table}{Table}{Tables}

\makeatletter
\DeclareRobustCommand\onedot{\futurelet\@let@token\@onedot}
\def\@onedot{\ifx\@let@token.\else.\null\fi\xspace}

\def\eg{\emph{e.g}\onedot} 
\def\ie{\emph{i.e}\onedot}

\def\etal{\emph{et al}\onedot}
\makeatother

\newcommand{\name}{WHAC\xspace}
\newcommand{\dataset}{WHAC-A-Mole\xspace}
\newcommand{\supp}{Supplementary Material\xspace}
\newcommand{\mv}{MotionVelocimeter\xspace}
\newcommand{\Sec}{Sec.\xspace}

\title{\name: World-grounded Humans and Cameras}

\author{
  Wanqi Yin\thanks{Equal contributions.}\hspace{4pt}$^{,1,2}$, Zhongang Cai$^{*,1,3,4}$, Ruisi Wang$^{1}$, Fanzhou Wang$^{1}$, \\
  \textbf{Chen Wei$^{1}$, Haiyi Mei$^{1}$, Weiye Xiao$^{1}$, Zhitao Yang$^{1}$, Qingping Sun$^{1}$,} \\
  \textbf{Atsushi Yamashita$^{2}$, Ziwei Liu$^{3}$, Lei Yang$^{1,4}$} \\
  $^{1}$ SenseTime Research, 
  $^{2}$ The University of Tokyo, \\
  $^{3}$ S-Lab, Nanyang Technological University,
  $^{4}$ Shanghai AI Laboratory
}

\begin{document}

\maketitle

\begin{abstract}

Estimating human and camera trajectories with accurate scale in the world coordinate system from a monocular video is a highly desirable yet challenging and ill-posed problem. In this study, we aim to recover expressive parametric human models (\ie, SMPL-X) and corresponding camera poses jointly, by leveraging the synergy between three critical players: the world, the human, and the camera. Our approach is founded on two key observations. Firstly, camera-frame SMPL-X estimation methods readily recover absolute human depth. Secondly, human motions inherently provide absolute spatial cues. By integrating these insights, we introduce a novel framework, referred to as \textbf{\name}, to facilitate world-grounded expressive human pose and shape estimation (EHPS) alongside camera pose estimation, without relying on traditional optimization techniques. Additionally, we present a new synthetic dataset, \textbf{\dataset}, which includes accurately annotated humans and cameras, and features diverse interactive human motions as well as realistic camera trajectories. Extensive experiments on both standard and newly established benchmarks highlight the superiority and efficacy of our framework. We will make the code and dataset publicly available.
  
\end{abstract}


\section{Introduction}
\label{sec:intro}

Expressive human pose and shape estimation (EHPS) has garnered considerable research attention due to its wide applications across the entertainment, fashion, and healthcare industries. Despite remarkable advancements in recent years, the majority of EHPS methods primarily focus on estimating parametric human models (\ie, SMPL-X~\cite{pavlakos2019expressive}) in the camera coordinate system. This approach falls short in dynamic situations where the camera and subject move concurrently. Estimating 3D trajectories in the world coordinate system (world-grounded) from 2D camera footage is challenging as the 3D-to-2D projection results in a loss of critical spatial information. Consequently, camera trajectories deduced are thus inherently ``scaleless", and the depth of humans directly estimated from the camera perspective lacks validity.

\begin{figure}[t]
  \centering
  \includegraphics[width=\textwidth]{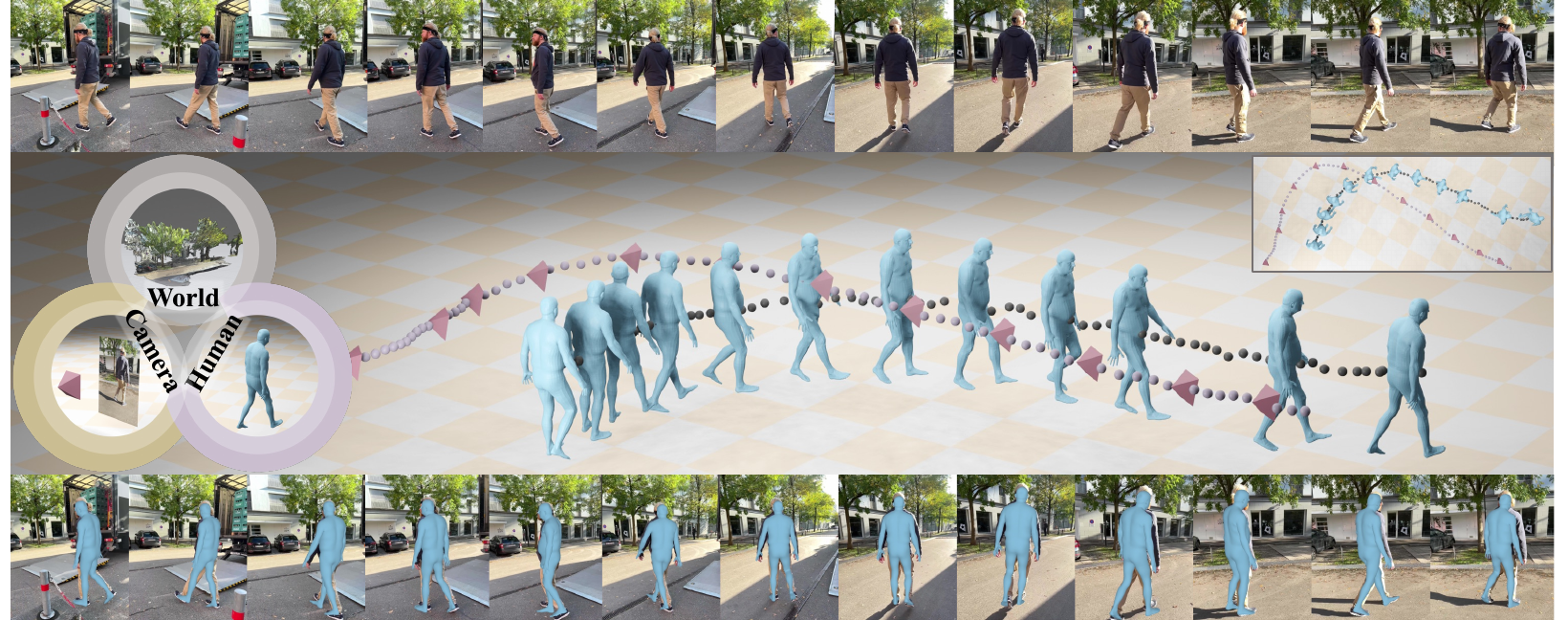}
  \vspace{-6mm}
  \caption{\name synergizes human-camera (camera-frame SMPL-X estimation), camera-world (visual odometry), and human-world (our proposed \mv) modeling for constructing world-grounded human and camera trajectories.}
  \label{fig:whac_teaser}
  \vspace{-4mm}
\end{figure}

In this work, we demonstrate the synergy between humans, cameras, and the world. First, existing camera-frame EHPS methods, although not specifically supervised to estimate human depth directly, can still accurately deduce the true depth. This only requires a reasonably accurate focal length that can be obtained from the video capture devices or estimated~\cite{kissos2020beyond}. Second, root translation is a critical component of human motions, allowing the latter to serve as a strong prior after an association is learned. Hence, by analyzing human poses, one can make an informed estimation of the velocity of human movement. Building upon these insights, we present \name, a novel framework designed to jointly estimate expressive human models and camera movements using a monocular video. For any given input video, camera-frame SMPL-X parameters and a preliminary camera trajectory are first estimated using plug-and-play EHPS~\cite{cai2023smpler} and visual odometry~\cite{teed2024deep} models. The human-camera relative positions are first deduced. These estimations are then utilized with VO estimations to canonicalize the sequences of human poses for accurate velocity estimation. Consequently, the scale of the camera trajectory can be recovered. It is noteworthy that \name pioneers whole-body, optimization-free estimation in a world-grounded context to recover human and camera trajectories jointly.

Moreover, the development of a new dataset becomes essential to more accurately assess model performance on world-grounded human motions and camera trajectories across a broader spectrum of scenarios. Recent studies have underscored the surprising efficacy of synthetic data~\cite{cai2023smpler, black2023bedlam, yang2023synbody, cai2021playing}, thanks to its diversity and controllability. Inspired by these findings, we introduce \dataset, a comprehensive synthetic dataset for \underline{W}orld-grounded \underline{H}umans \underline{A}nd \underline{C}ameras with a rich collection of \underline{A}nimated subjects under \underline{MO}ving viewpoints in mu\underline{L}tiple \underline{E}nvironments. \dataset features comprehensive motion sequences that include 1) interactive human activities from DLP-MoCap~\cite{cai2023digital}, 2) partner dances from DD100~\cite{siyao2023duolando}, in addition to 3) the standard AMASS~\cite{mahmood2019amass} motion repository. Notably, \dataset includes automatically generated camera trajectories that mimic cinematic filming techniques, such as \textit{tracking shots} and \textit{arc shots}, thereby offering a high level of realism. 

We validate our \name on standard benchmarks and \dataset to obtain consistent performance gains compared to the state-of-the-art (SoTA) methods under both camera-frame and world-grounded settings. \name even demonstrates a surprising capability to handle corner cases when motion-based and camera-based observations contradict, paving the way for potential applications.

In summary, our contributions are three-fold. First, we propose \name, a novel regression-based framework that capitalizes on human priors for the pioneering world-grounded EHPS method. Second, we contribute \dataset, a comprehensive benchmark with accurate human and camera annotations of diverse human activities. Third, our empirical evaluations underscore the superior performance of \name across multiple benchmarks.

\section{Related Works}
\label{sec:related_works}

\subsection{Expressive Human Pose and Shape Estimation (EHPS)}

EHPS captures body, face, and hands from monocular images or videos, typically through parametric human models (\eg, SMPL-X~\cite{pavlakos2019expressive}). Early optimization-based method~\cite{pavlakos2019expressive} fits SMPL-X models on 2D keypoints, and was soon outperformed by regression-based methods that were trained on a large amount of paired data. Two-stage methods estimate body parameters first, then hand/face parameters from crop-out image patches~\cite{choutas2020monocular, rong2021frankmocap, zhou2021monocular, feng2021collaborative, moon2022accurate, li2023hybrik, zhang2023pymaf, pang2024towards}. Recently, OSX~\cite{lin2023one} proposes the one-stage paradigm that estimates body, hand, and face with shared features. This paradigm shift simplifies the pipeline and led to the first foundation model SMPLer-X~\cite{cai2023smpler} that achieved unprecedented generalization ability across key benchmarks. However, despite their success, these methods estimate parametric humans in the camera coordinate, lacking information on the global trajectory especially when the camera is moving.

\subsection{World-grounded Recovery of Humans and Cameras}

Estimation of human trajectory in world coordinate system typically requires a multi-camera setup~\cite{hasler2009markerless, joo2015panoptic, zhang20204d, peng2021neural, huang2021dynamic, cai2022humman, renbody} or additional wearable devices (\eg., IMU~\cite{von2018recovering, guzov2021human} or electromagnetic sensors~\cite{kaufmann2023emdb}). Methods that require only a single camera often rely on other assumptions: Yu \etal~\cite{yu2021human} needs the scene to be provided by the user and Luvizon \etal~\cite{luvizon2023scene} assumes a static camera. D\&D~\cite{li2022d}, GLAMR~\cite{yuan2022glamr}, and TRACE~\cite{sun2023trace} estimate global human trajectories from single-frame poses or image features. However, camera and human rotation have a coupled effect on camera-frame global orientation estimation, which leads to ambiguity. Liu \etal~\cite{liu20214d} leverages Structure-from-Motion (SfM)~\cite{schonberger2016structure} to reconstruct both camera and human trajectories and adjust human's scale to match the camera's, which may not reflect the absolute scale. Recently, SLAHMR leverages SLAM~\cite{teed2021droid} and human motion prior~\cite{rempe2021humor} in the optimization to recover humans and the camera. However, the process is computationally expensive and takes excessively long to complete. PACE~\cite{kocabas2023pace} also leverages visual odometry~\cite{teed2024deep} for camera pose estimation and a faster human motion to significantly accelerate the optimization process but is still time-consuming. WHAM~\cite{shin2023wham} is the first regression-based work in the domain that features real-time performance. It takes camera estimation (angular velocity) as the input and estimates human parameters in the camera frame and human trajectory in the world frame through separate branches, while the camera trajectory is not recovered. Our \name aims to recover both human and camera trajectories in the world coordinate with accurate scales.
\section{Methodology}
\label{sec:method}
\begin{figure}[t]
  \centering
  \includegraphics[width=\textwidth]{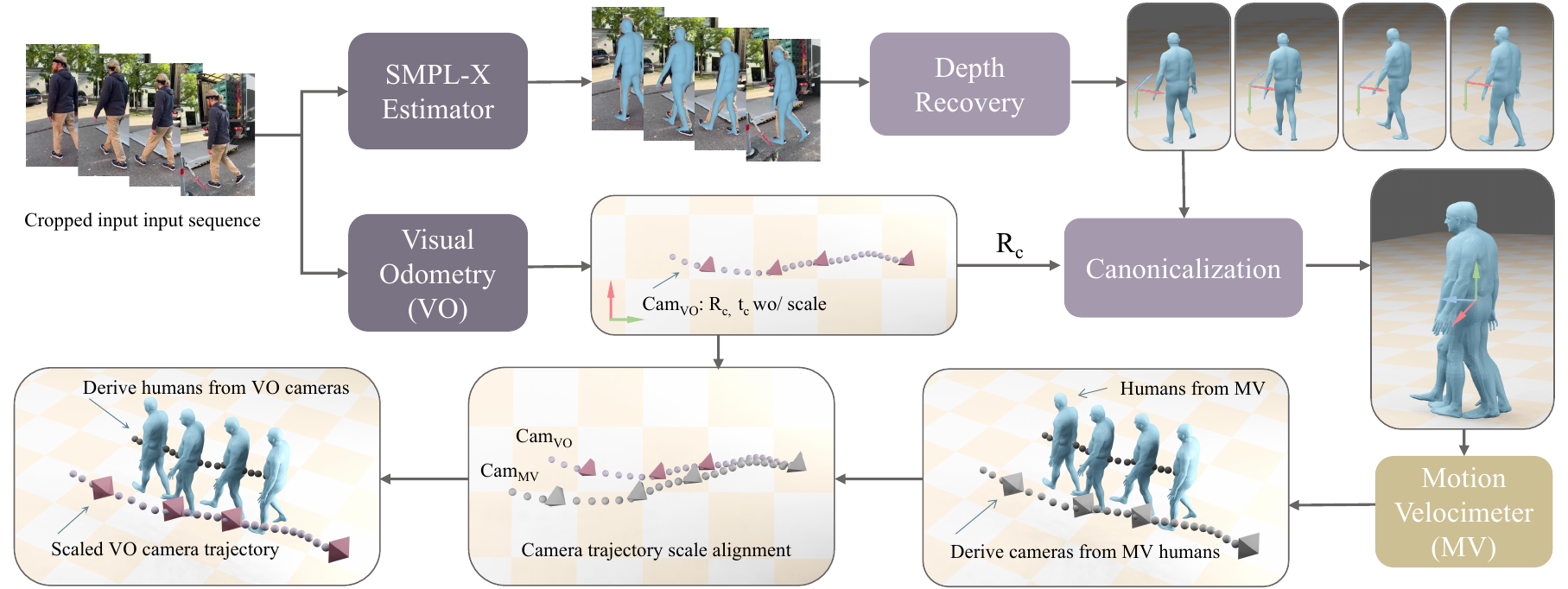}
  \caption{Overview of \textbf{\name}. SMPL-X estimator extracts camera-frame SMPL-X with dummy depth, which is recovered in \cref{sec:method:recover_depth}. The scaleless camera trajectory estimated by VO is then used to canonicalize the human trajectory to estimate its velocity and thus scale in \cref{sec:method:recover_scale}. A camera trajectory is then derived for alignment and scale recovery, which subsequently updates the human trajectory in \cref{sec:method:recover_trajectories}.}
  \label{fig:method}
  \vspace{-4mm}
\end{figure}

Recovering accurate 3D dimensions from 2D observations is an ill-posed problem: a small object at a close range may appear the same as a large object at a far range. In this section, we aim to address two ambiguities with priors that are surprisingly effective, but not thoroughly utilized in existing EHPS works: the parametric humans themselves. 

\subsection{Preliminaries}

\paragraph{Problem Formulation.} We aim to estimate human and camera pose sequences in the world coordinate system. The humans are represented by SMPL-X parameters: global orientation $\theta^{w}_{go} \in \mathbb{R}^{1\times3}$, translation $t^{w}_{h} \in \mathbb{R}^{1\times3}$, body pose $\theta_{b} \in \mathbb{R}^{21\times3}$, left hand pose $\theta_{lh} \in \mathbb{R}^{15\times3}$, right hand pose $\theta_{rh} \in \mathbb{R}^{15\times3}$, jaw pose $\theta_{j} \in \mathbb{R}^{1\times3}$, body shape $\beta \in \mathbb{R}^{10}$ and facial expression $\phi \in \mathbb{R}^{10}$. The cameras are represented by world-frame rotation $R^{w}_{c} \in \mathbb{R}^{1\times3}$ and translation $t^{w}_{c} \in \mathbb{R}^{1\times3}$. In this work, the superscript indicates the coordinate system (\ie, $w$ for world and $c$ for camera).

\paragraph{Camera-frame SMPL-X Estimation} typically omits absolute depth estimation. Hence, primary metrics (\eg, PA-MPJPE, MPJPE, and PVE) all perform root alignment. We employ SMPLer-X~\cite{cai2023smpler}, a strong foundation model that demonstrates accurate estimation of human pose and shapes. We add additional GRUs before the prediction heads and finetune the model to better capture the temporal cues.

\paragraph{Visual Odometry (VO)} typically provides high-quality $R^{w}_{c}$; the trajectory formed by $t^{w}_{c}$ is scaleless but accurate in shape. We also follow the standard to define the first camera frame of the input video as the world coordinate system.

\subsection{Recovering Camera-space Human Root Translation}
\label{sec:method:recover_depth}

\begin{figure}[t]
  \centering
  \includegraphics[width=0.75\textwidth]{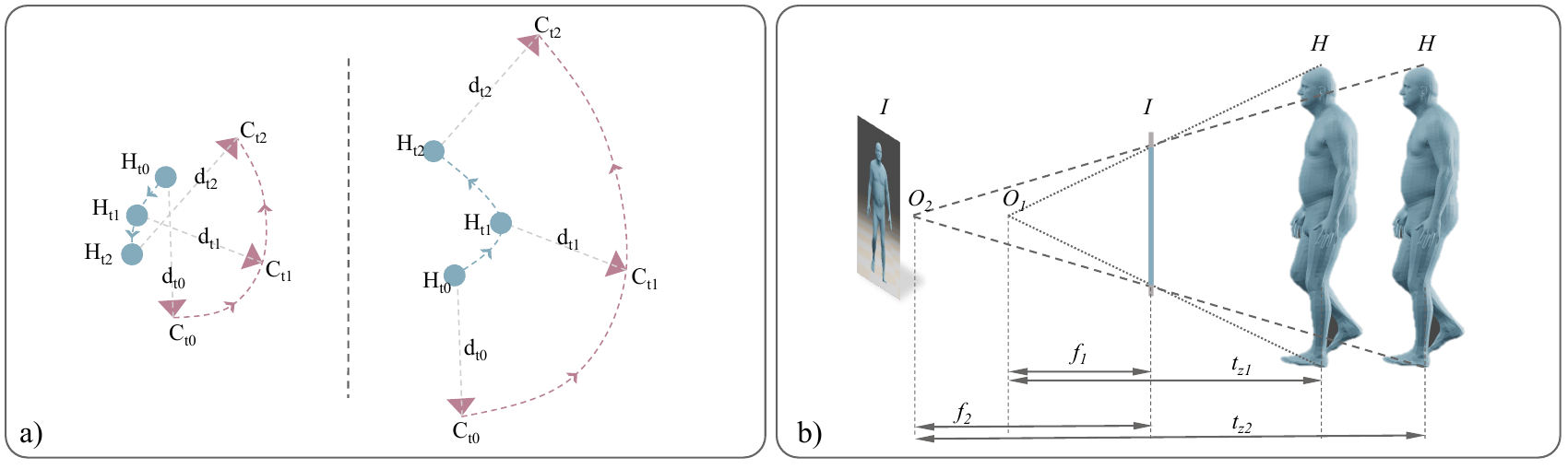}
  \vspace{-2mm}
  \caption{a) Human trajectories $H$ derived from camera trajectories $C$ of different scales can be vastly different in both shape and direction, despite that the same camera-frame human root depth $d_{t}$ and translations $t^{c}_{h}$ are used. b) Different pairs of focal length $f$ and $t_z$ can correspond to the same image.}
  \label{fig:camera_illustration}
  \vspace{-2mm}
\end{figure}

Mainstream EHPS methods~\cite{moon2022accurate, lin2023one, cai2023smpler} recover parametric humans in the camera space and adopt a weak perspective camera model, which considers all points to be at the same depth away from the camera.
\begin{equation}
    \begin{bmatrix}
    f^{*} & 0 & 0 \\
    0 & f^{*} & 0 \\
    0 & 0 & 1
    \end{bmatrix}
    \begin{bmatrix}
    t_{x} + {\delta}x \\
    t_{y} + {\delta}y \\
    t_{z} + {\delta}z
    \end{bmatrix}
    \approx
    \begin{bmatrix}
    f^{*} & 0 & 0 \\
    0 & f^{*} & 0 \\
    0 & 0 & 1
    \end{bmatrix}
    \begin{bmatrix}
    t_{x} + {\delta}x \\
    t_{y} + {\delta}y \\
    t_{z}
    \end{bmatrix}
\end{equation}
where $f^{*}$ indicates the focal length in NDC (Normalized Device Coordinate) space (image pixel coordinates are normalized into $[-1, 1]$). $t^{c}_h = (t_x, t_y, t_z)$ is the root translation in the camera frame, $({\delta}x, {\delta}y, {\delta}z)$ is the relative translation of a point relative to the root. Hence, the projected 2D point (NDC space) is written as:

\begin{equation}
    \begin{bmatrix}
    u^{*}  \\
    v^{*} 
    \end{bmatrix}
    =
    \begin{bmatrix}
    f^{*}(t_{x} + {\delta}x)/t_{z} \\
    f^{*}(t_{y} + {\delta}y)/t_{z}
    \end{bmatrix}
    =
    \begin{bmatrix}
    s(t_{x} + {\delta}x) \\
    s(t_{y} + {\delta}y)
    \end{bmatrix}
\end{equation}
where $s$ is the scale parameter. The SMPL-X estimator also predicts camera parameters $(s, t_x, t_y)$ to reproject SMPL-X joints on the image plane. Hence, we obtain the relationships between focal length, depth, and $s$:
\begin{equation} \label{eqn:s}
    s = \frac{f^{*}}{t_{z}} = \frac{2}{I} \times \frac{f}{t_{z}} \Rightarrow t_{z} = \frac{2}{I} \times \frac{f}{s}
\end{equation}
where $f$ is the focal length in pixels. $I$ is the resolution (side pixel length) of the input crop to the SMPL-X estimator. 

We point out that although these camera-frame methods do not supervise the human root depth $t_{z}$, by training the model to produce a scale $s$ that overlays SMPL-X accurately back on the image plane, the model implicitly learns human root depth $t_{z}$ that is coupled with focal length $f$ as illustrated in \cref{fig:camera_illustration}b). A dummy focal length of 5,000 is often used~\cite{moon2022accurate, lin2023one, cai2023smpler}, however, this leads to unrealistic human root depth $t_{z}$. We highlight that accurate intrinsic parameters are accessible from many devices, and our empirical results show using the diagonal pixel length~\cite{kissos2020beyond} also yields satisfactory results.

\subsection{Estimating World-frame Human Motions for Scale Extraction}
\label{sec:method:recover_scale}

In recent years, there has been a plethora of high-quality optical motion capture datasets that become available, covering a wide range of human activities. Previous art~\cite{shin2023wham} estimates human global trajectory from 2D keypoint observations, which may not capture subtle 3D information. Hence, we propose to learn absolute scale from 3D human motions. 

First, for a SMPL-X sequence of $K$ frames, estimated in the camera coordinate system, we compute the 3D joint coordinates. Specifically, for each frame:
\begin{equation}
    J^{c} = M(\theta^{c}_{go}, t^{c}, \theta_{b}, \theta_{lh}, \theta_{rh}, \theta_{j}, \beta, \phi), J^{c} \in \mathbb{R}^{K\times15\times3}
\end{equation}
where $M$ is the SMPL-X parametric model. We select 15 joints (14 LSP~\cite{johnson2010clustered} joints and the pelvis) from the original 55 joints. We then compute the joints in the world coordinate system:
\begin{equation}
    J^{w} = T^{vo} \times J^{c}, T^{vo} = [R^{w}_{c} | t^{w}_{c}]
\end{equation}
where $T^{vo}$ is the visual odometry's estimation (camera-to-world transformation). Note that $t^{w}_{c}$ does not have a valid scale. 

To facilitate the training, we standardize the input: we define a canonical frame where the human is root-aligned with zero global orientation. We then compute the canonical transformation $T^{cano}$ by using the first ($0^{th}$) frame's rotation and offset the translation to zero:
\begin{equation}
    T^{cano} = [R^{cano} | t^{cano}], 
    R^{cano} = (R^{w}_{c,0} \times \theta^{c}_{go,0})^{-1} = (\theta^{c}_{go,0})^{-1},
    t^{cano} = - p^{w}_{0}
\end{equation}
where $R^{w}_{c,0}$ is $0^{th}$ camera rotation in the world frame estimated from visual odometry, $\theta^{c}_{go,0}$ is $0^{th}$ global orientation estimated in the camera space, $p^{w}_{0}$ is the pelvis joint of $J^{w}_{0}$. Note $R^{w}_{c,0}$ is an identity matrix $I_{3}$ as the $0^{th}$ camera frame is defined as the world frame. All joints are then canonicalized as $J^{cano} = T^{cano} \times J^{w}$.

Our \mv then estimates per-frame velocity in the canonical space:
\begin{equation}
    V^{cano} = \text{\mv}(J^{cano})
\end{equation}
where the velocity is then de-canonicalized back to the world frame:
\begin{equation}
    V^{world} = (T^{cano})^{-1} \times V^{cano}
\end{equation}
with $V^{world}$, we can reconstruct the human trajectory with scale in the world coordinate system. \mv only requires a simple architecture that we include in the \supp.

\subsection{Recovering Scaled Human and Camera Trajectories}
\label{sec:method:recover_trajectories}

As we obtain human trajectory $t^{w}_h$ with absolute scale, one possible way is to align the human trajectory derived from the VO-estimated camera trajectory using camera-frame human root translation to $t^{w}_h$. However, \cref{fig:camera_illustration}a) shows that such alignment is problematic as the human trajectory derived from scaleless camera trajectory may be invalid. Hence, we propose to transfer the scale to the camera trajectory in two steps. First, we derive a camera trajectory from the human trajectory:
\begin{equation}
    T^{w}_{c,derived} =(T^{cano})^{-1} \times  T^{cano}_{h} \times (T^{c}_{h})^{-1}
\end{equation}

\begin{equation}
    T^{w}_{c,derived}=[R^{w}_{c,derived} | t^{w}_{c,derived}]
\end{equation}
This derived camera trajectory already has an accurate scale with a good shape. However, we find that the camera trajectory estimated by VO has a better, more robust shape because it can leverage visual cues that are much denser than human motion cues. In this light, we perform Umeyama's method~\cite{umeyama1991least} (shown as $\overset{U}{\rightarrow}$) to align the VO-estimated camera trajectory with the human-derived camera trajectory $t^{w}_{c,derived}$ while discard $R^{w}_{c,derived}$ and keeping the camera rotation $R^{w}_{c}$:
\begin{equation}
    t^{w}_{c,final} = t^{w}_{c} \overset{U}{\rightarrow} t^{w}_{c,derived}
\end{equation}
Hence, we then update human trajectory by deriving it from the aligned camera trajectory $t^{w}_{c,final}$:
\begin{equation}
    T^{w}_{h,final} = T^{w}_{c,final} \times T^{c}_h,T^{w}_{c,final}=[R^{w}_{c} | t^{w}_{c,final}],T^{w}_{h,final}=[R^{w}_{h,final} | t^{w}_{h,final}]
\end{equation}
As a result, we obtain human trajectory $t^{w}_{h,final}$ and camera trajectory $t^{w}_{c,final}$, both in the world coordinate system and with absolute scales.


\section{\dataset Dataset}
\label{sec:dataset}

\setlength{\tabcolsep}{4pt}

\begin{table}[t]
  \caption{\textbf{Dataset Comparison.} \#Inst.: number of human instances (crops). \#Seq.: number of video sequences. R/S: Real or Synthetic. Multi.: multiperson scenes. Track.: track ID labels. HHI: human-human interaction motions. $\dagger$: EgoSet. $\ddagger$: unknown as the data is not released when this paper is written. $\diamond$: typically short (<100 frames) clips. }
  \label{tab:dataset_comaprison}
  \centering
  \begin{tabular}{@{}lccccccccccc@{}}
    \toprule
        Dataset & 
        \#Inst. & \#Seq. & R/S & Multi. & Track. & Contact & HHI & Camera & Human \\
    \midrule

        3DPW~\cite{von2018recovering} &           
        74.6K  & 60 &  R & \checkmark & $\times$ & $\times$ & \checkmark & Moving & SMPL  \\ 
  
        RICH~\cite{huang2022capturing}  &          
        476K  & 141 & R & \checkmark & \checkmark & \checkmark & \checkmark & Static* &  SMPL \\          
        
        HCM~\cite{kocabas2023pace} &
        $\ddagger$  & 25 & S & \checkmark & $\ddagger$ & $\times$ & $\times$ & Moving & SMPL \\           
        EMDB~\cite{kaufmann2023emdb} &        
        109K & 81 & R & $\times$ & N.A. & $\times$ & N.A. & Moving & SMPL \\   

        EgoBody$^{\dagger}$~\cite{zhang2022egobody}  &
        175K  & 125 & R & \checkmark & \checkmark & $\times$ & \checkmark & Moving & SMPL-X  \\  
        BEDLAM~\cite{black2023bedlam} &          
        951K  & 10.4K$^{\diamond}$ & S & \checkmark & \checkmark & $\times$ & $\times$ & Static & SMPL-X \\ 

        SynBody~\cite{yang2023synbody} &       
        2.7M & 27K$^{\diamond}$ & S & \checkmark & \checkmark & $\times$ & $\times$ & Static & SMPL-X \\  
        
    \midrule
        \dataset &       
        1.46M & 2434 & S & \checkmark & \checkmark & \checkmark & \checkmark & Moving  & SMPL-X \\ 
        
  \bottomrule
  \end{tabular}
  \vspace{-4mm}
\end{table}

\begin{figure}[t]
  \centering
  \includegraphics[width=\textwidth]{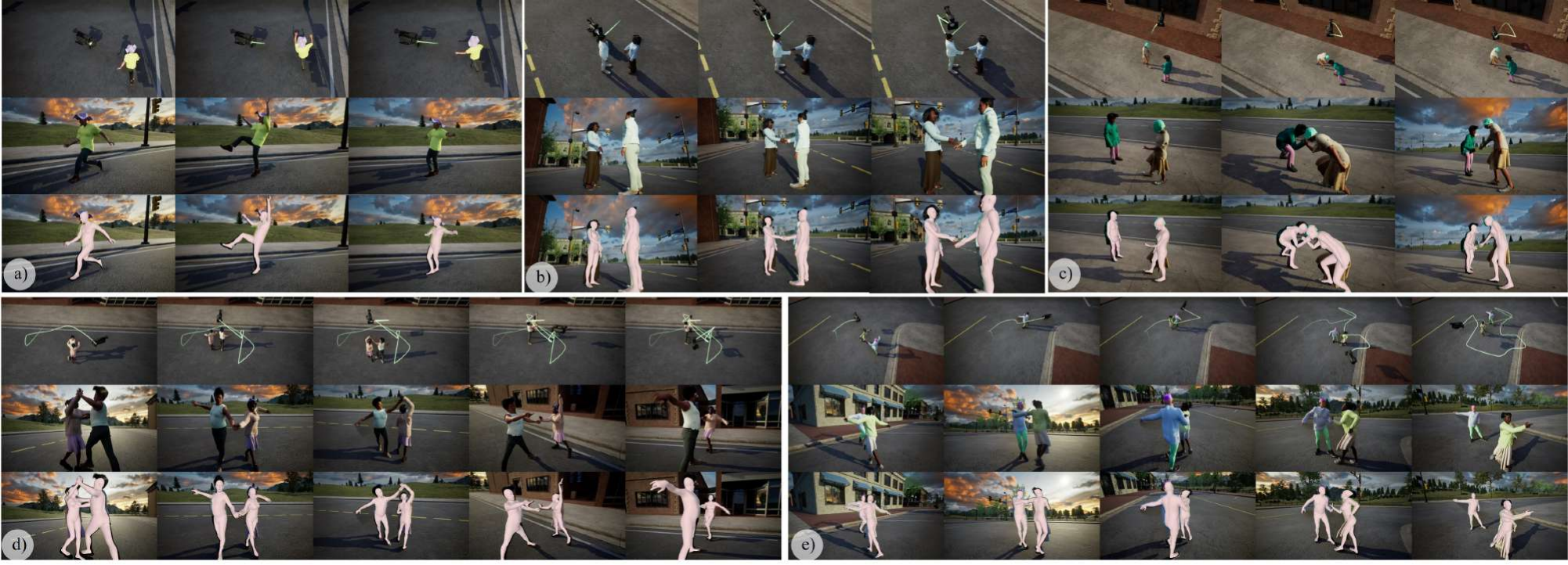}
  \vspace{-4mm}
  \caption{Visualization of \dataset sample sequences, animated with a) AMASS, b-c) DLP-MoCap, and d-e) DD100. In each sample, the first row depicts the overview (note the camera trajectory shown in bright rays), and the second and the third rows show the camera view and overlaid SMPL-X annotations. }
  \label{fig:whac_a_mole}
\end{figure}

We highlight that \dataset combines fine-crafted automatic camera movements with varied characters animated with diverse, high-quality motion sequences to generate a dataset with accurate camera and SMPL-X annotations. The dataset is constructed with the advanced human data synthesis toolbox XRFeitoria~\cite{xrfeitoria}. It leverages SMPL-XL (a layered extension of SMPL-X) to create virtual humans with diverse body shapes, clothing, and accessories. We follow SynBody~\cite{yang2023synbody} in the scene setup, subject creation, and placement. We further improve the data synthesized in two ways: diverse motion sources (\cref{sec:dataset:interative_human_motions}) and camera trajectory generation (\cref{sec:dataset:camera_trajectory_generation}).
In \cref{tab:dataset_comaprison}, we compare \dataset with popular video-based benchmarks with both camera and human annotations. \dataset features a competitive scale of training instances and video sequences, multiperson scenes with track IDs, contact labels, accurate camera pose and SMPL-X annotations. We split \dataset by motion sequence into 80\%:20\% for training and testing.
Examples of \dataset are visualized in \cref{fig:whac_a_mole}.

\subsection{Interactive Human Motions}
\label{sec:dataset:interative_human_motions}

AMASS~\cite{mahmood2019amass} is a popular motion repository, widely used by existing synthetic datasets~\cite{kocabas2023pace, black2023bedlam, yang2023synbody}. However, AMASS only contains single-person motions. As a result, synthetic data is captured in virtual scenes populated with unrelated single-person motions, typically scattered sparsely to avoid collision. However, close human interactions are common in daily life, and difficult to solve. In this light, we select two latest motion datasets that contain comprehensive interactive human motions. First, DD100~\cite{siyao2023duolando}, a duet dance motion capture dataset that includes near two hours of partner dances of 10 different genres. Second, DLP-MoCap~\cite{cai2023digital}, a motion capture dataset containing daily interactions between two subjects. Since SMPL-XL models are fully compatible with SMPL-X body pose sequences, we animate virtual characters with a combination of AMASS, DD100, and DLP-MoCap.

\subsection{Camera Trajectory Generation}
\label{sec:dataset:camera_trajectory_generation}

To better model the camera movement, we adopt the representation in Rao \etal~\cite{anyi2023dynamic} to define the camera in a human-centric spherical coordinate system \((r_c, \theta_c, \phi_c)\), in which the \(r_c\) represents the distance from the camera to the character, while the polar angle \(\theta_c\) and the azimuthal angle \(\phi_c\) define the angle between the camera's looking direction and the character's facing direction. Therefore, given a character's location \((x_{ch}, y_{ch}, z_{ch})\) and facing direction \((\theta_{ch}, \phi_{ch})\), the camera's location in the world space is
\begin{equation}
 (x_c, y_c, z_c) = (x_{ch}, y_{ch}, z_{ch}) + r_c(\sin(\theta)\cos(\phi), \sin(\theta)\sin(\phi),\cos(\theta))
\end{equation}
where the \(\theta = (\theta_c + \theta_{ch}) \mod{2\pi}\), the \(\phi = (\phi_c + \phi_{ch}) \mod{2\pi}\), and the camera's rotation is thereby calculated by restricting the camera look at the \((x_{ch}, y_{ch}, z_{ch})\). In \dataset, we design two types of shot scales including the \textit{medium shot} and the \textit{full shot}, which respectively use the location of the \textit{neck} and the \textit{pelvis} as the character's location \((x_{ch}, y_{ch}, z_{ch})\). 
For the motion sequences that consist of multiple characters, the \((x_{ch}, y_{ch}, z_{ch})\) and the \((\theta_{ch}, \phi_{ch})\) are derived from the average of the locations and the facing directions of all the characters.
Based on the human-centric spherical coordinate system \((r_c, \theta_c, \phi_c)\), we design different keyframe-setting strategies to simulate five common camera movements below.

\noindent \textbf{- Arc shot} adds equally-spaced keyframes to rotate the camera around the character horizontally by increasing or decreasing \(\phi_c\in[\phi_{min}, \phi_{max}] \) or vertically by increasing or decreasing \(\theta_c\in[\theta_{min}, \theta_{max}]\). Moreover, the angular velocity of the \textit{arc shots} can be controlled by adjusting the \(\Delta\phi_c\) or the \(\Delta\theta_c\) between two adjacent keyframes.

\noindent \textbf{- Push shot} also adds equally-spaced keyframes and moves the camera towards the character by decreasing the \(r_c\). More specifically, the fraction of the character in the view, which is more intuitive when filming, is used to indirectly decrease the \(r_c\) by
\begin{equation}
    r_c = \left\{
    \begin{array}{cc}
        \frac{ h_{bbox} } { frac * \tan({fov/2})} \quad as \leqslant 1.0 \\
        \frac{ h_{bbox}  * as } { frac * \tan({fov/2})} \quad as > 1.0
    \end{array}
    \right.
\end{equation}
where the \(h_{bbox}\) is the height of the character's bounding box in the camera space, the \(frac\) is the desired fraction of the character in the view, the \(as\) is the aspect ratio of the camera frame, and the \(fov\) is the field of view of the camera. Similar to the \textit{arc shots}, the speed of the camera movement can be adjusted by setting the \(\Delta frac\) between two adjacent keyframes.

\noindent \textbf{- Pull shot} is opposite to the \textit{push shot} and moves the camera further away from the character by increasing the \(r_c\) under the control of the \(frac\). Randomly sampling \(frac\) in a range \([frac_{min}, frac_{max}]\) at different keyframes derives continuous pushing and pulling, which is commonly used when filming dances.

\noindent \textbf{- Tracking shot} follows the character and maintains the relative position between the camera and the character, \ie, maintain the \((r_c, \theta_c, \phi_c)\) of the camera in the human-centric spherical coordinate system. A new keyframe of the \textit{tracking shot} is added when the overlap ratio of the character's bounding box in the current frame and in the last keyframe is greater than a threshold \(\lambda_{overlap}\).

\noindent \textbf{- Pan shot} rotates the camera horizontally to keep the camera looking at the character, therefore it is another way to make the camera follow the character, and it shares the same rule with the \textit{tracking shot} to add a new keyframe. 

Rather than assigning a specific camera movement to an entire human motion sequence, our pipeline automatically combines several types of camera movements into one motion sequence to increase the variety of camera movements. For example, when capturing static motions (whose longest edge of the bounding box formed by the \((x_{ch}, y_{ch}, z_{ch})\) across all frames is less than a threshold \(\lambda_{bbox}\)) or interactive motions, we combine the horizontal and the vertical \textit{arc shots} with the random \textit{pull} or \textit{push shots} to rotate the camera around the characters as well as transiting smoothly between different shot angles, such as high-angle, low-angle or eye-level, and pushing in or pulling out the distance between the camera and the characters to increase the rhythm of the camera movement. For the motions with long-distance movements, we combine the \textit{tracking shots} and the \textit{pan shots} to follow the character. If the character's facing direction is stable (\ie, that the rotation angle from the character's facing direction in the last keyframe to the current keyframe is less than a threshold \(\lambda_{angle}\)), we use the  \textit{tracking shot}. Otherwise, we use the  \textit{pan shot}. This rule effectively smooths the camera's movement, especially when the character turns dramatically.

\section{Experiments}
\label{sec:experiments}
We evaluate \name on both camera-frame and world-grounded benchmarks to compare its parametric human recovery abilities with existing SoTA methods. Due to space constraints, we include inference speed comparison, more visualizations on trajectory reconstruction, and more qualitative results in the \supp.

\subsection{Implementation Details}
We finetune SMPLer-X-B~\cite{cai2023smpler} with EgoBody, 3DPW, and EMDB for camera-frame estimation of SMPL-X parameters. \dataset (with motions from AMASS, DD100, and DLP-MoCap), 3DPW, EMDB, and RICH are used to train the MotionVelocimeter. More details are in the \supp.

\subsection{Datasets}
In addition to our proposed \dataset, mainstream benchmarks for human pose and shape estimation with parametric human labels are used. \textbf{EgoBody}~\cite{zhang2022egobody} includes 125 sequences of 36 subjects in 15 indoor scenes, featuring 3D human motions interacting with scenes. We study the EgoSet that is captured by a head-mounted camera; 2) \textbf{3DPW}~\cite{von2018recovering}, a popular dataset with 60 sequences captured by an iPhone, featuring diverse human activities in outdoor scenes; 3) \textbf{EMDB}~\cite{kaufmann2023emdb} provides 58 minutes of motion data of 10 subjects in 81 indoor and outdoor scenes. Notably, it contains a subset, EMDB 2, that contains global trajectories of humans and cameras. 4) \textbf{RICH}~\cite{huang2022capturing} consists of 142 multi-view videos with 22 subjects and 5 scenes with 6-8 fixed cameras. RICH is not used for evaluation as the cameras are static.

\subsection{Evaluation Metrics}

\paragraph{For camera-frame human recovery,} we use the standard Mean Per Joint Position Error (\textbf{MPJPE}), Procrustes-aligned MPJPE (\textbf{PA-MPJPE}), Per Vertex Error (\textbf{PVE}) in millimeters (mm), and Acceleration error (\textbf{Accl.}) in m/s$^{2}$. Note that these metrics are evaluated after root alignment between estimated and ground truth parametric humans, thus not considering discrepancy in translation estimation. In this light, we also report \textbf{T-MPJPE}~\cite{bazavan2021hspace} and similarly \textbf{T-PVE}, which are variants of MPJPE and PVE that includes translation estimation to reflect the accuracy of depth estimation in the camera space. 

\paragraph{For world-frame human/camera recovery,} we follow previous works~\cite{ye2023decoupling, kocabas2023pace, shin2023wham} to split human motion sequences with global trajectory into 100-frame segments. The segments are Procrustes-aligned to the ground truth for MPJPE computation: \textbf{W-MPJPE$_{100}$} if the first two frames are used in the alignment or \textbf{WA-MPJPE$_{100}$} if the entire segment. To evaluate the quality of trajectory, we extend Average Trajectory Error (ATE)~\cite{kocabas2023pace} to \textbf{C-ATE} and \textbf{H-ATE} for camera and human respectively, which are computed after Procrustes-alignment of estimated and ground truth trajectories. All metrics are in millimeters (mm). We also report respective Alignment Scales (AS) used in the alignment for the camera (\textbf{C-AS}) and human (\textbf{H-AS}) and values closer to 1.0 indicate more accurate scale estimation.

\subsection{World-grounded Benchmarks}
\setlength{\tabcolsep}{4pt}

\begin{table}[t]
  \caption{World-frame evaluation on \textbf{\dataset}. *: adapted to world-grounded evaluation. H-AS and C-AS: the closer to 1.0, the better.}
  \label{tab:world_space_whac_a_mole}
  \centering
  \resizebox{\textwidth}{!}{
  \begin{tabular}{@{}lrrrrrrrr@{}}
    \toprule
        & PA-MPJPE$\downarrow$ & W-MPJPE$\downarrow$ & WA-MPJPE$\downarrow$ & H-ATE$\downarrow$ & H-AS & C-ATE$\downarrow$ & C-AS \\
    \midrule
        OSX*~\cite{lin2023one} + DPVO~\cite{teed2024deep}          & 90.1 & 1036.1 & 390.7 & 180.5 & 0.5 & 0.5 &  7.3  \\
        SMPLer-X-B*~\cite{cai2023smpler} + DPVO~\cite{teed2024deep}   & 76.7 & 842.3 & 335.4 & 138.3 & 0.5 & 0.5 & 7.3  \\
        \name (GT Gyro)      & 76.5 & 343.8 & 182.0 & \textbf{103.5} & \textbf{0.9} & \textbf{0.5} &  \textbf{1.3}  \\
        \name          & \textbf{76.5} & \textbf{343.3} & \textbf{182.0} & \textbf{103.5} & \textbf{0.9} & \textbf{0.5} &  \textbf{1.3}  \\
    \bottomrule
  \end{tabular}}
\end{table}

\setlength{\tabcolsep}{4pt}

\begin{table}[tb]
  \vspace{-3mm}
  \caption{World-frame evaluation on \textbf{EMDB2}. *: adapted to world-grounded evaluation. H-AS and C-AS: the closer to 1.0, the better.}
  \label{tab:world_space_emdb2}
  \centering
  \resizebox{\textwidth}{!}{
  \begin{tabular}{@{}lrrrrrrrr@{}}
    \toprule
         & PA-MPJPE$\downarrow$ & W-MPJPE$\downarrow$ & WA-MPJPE$\downarrow$  & H-ATE$\downarrow$ & H-AS  & C-ATE$\downarrow$ & C-AS \\
    \midrule
        GLAMR~\cite{yuan2022glamr}               & 56.0 & 756.1 & 286.2 & - & - & - & -  \\
        SLAHMR~\cite{ye2023decoupling}              & 61.5 & 807.4 & 336.9 & 207.8 & 1.9 & - & -  \\
        WHAM~\cite{shin2023wham} (GT Gyro)      & 41.9 & 436.4 & 165.9 & 83.2 & 1.5 & - & -  \\
    \midrule
        OSX-L*~\cite{lin2023one} + DPVO~\cite{teed2024deep}    & 99.9 & 1186.2 & 458.8 & 235.4 & 2.3 & \textbf{14.8} & 5.1 \\
        SMPLer-X-B*~\cite{cai2023smpler} + DPVO~\cite{teed2024deep} & 42.5 & 930.1 & 375.8 & 200.6 & 2.0 & \textbf{14.8} & 5.1 \\
        \name (GT Gyro)     & \textbf{39.4} & 392.5 & 143.1 & \textbf{75.8} & \textbf{1.1} & \textbf{14.8} & 1.5   \\
        \name                & \textbf{39.4} & \textbf{389.4} & \textbf{142.2} & 76.7 & \textbf{1.1} & \textbf{14.8} & \textbf{1.4} \\
  \bottomrule
  \end{tabular}}
\end{table}

In \cref{tab:world_space_emdb2}, we evaluate on \dataset in \cref{tab:world_space_whac_a_mole}. \dataset provides expressive human (\ie, SMPL-X), with accurately annotated camera motions. Since no existing EHPS methods produce SMPL-X in the world coordinate system and a strictly fair comparison is not plausible, we build the first benchmark by making two adaptations to the SoTA methods (OSX~\cite{lin2023one} and SMPLer-X~\cite{cai2023smpler}): camera-frame translation estimation and visual odometry. Implementation details are included in the \supp. It is noted that methods that achieve good results on EMDB still struggle on \dataset, which can be attributed to the more challenging scenarios of \dataset (involving hard poses, diverse interactions, occlusions, and complicated camera movements). We hope \dataset can serve as a useful foundation for future world-grounded EHPS research.

Moreover, we compare \name with both body-only methods (GLAMR~\cite{yuan2022glamr}, SLAHMR~\cite{ye2023decoupling}, and WHAM~\cite{shin2023wham}) and whole-body methods (OSX-L~\cite{lin2023one} and SMPLer-X~\cite{cai2023smpler}) on EMDB2, where WHAC achieves best performance, even surpassing body-only methods that are native to EMDB's SMPL annotations.

\subsection{Camera-space Benchmarks}
\setlength{\tabcolsep}{4pt}

\begin{table}[tb]
  \caption{Results of camera-frame methods on \textbf{EgoBody (EgoSet)} with SMPL-X ground truths. PVE variants are measured for whole-body (SMPL-X) methods only.}
  \label{tab:cam_space_egobody}
  \centering
  \vspace{-3mm}
  \begin{tabular}{@{}lrrrrrrr@{}}
    \toprule
          & PA-MPJPE$\downarrow$ & PA-PVE-all$\downarrow$ & PVE-all$\downarrow$ & PVE-hand$\downarrow$ & PVE-face$\downarrow$  & Accl.$\downarrow$ \\
    \midrule
        GLAMR~\cite{yuan2022glamr} & 114.3 & - & - & - & - & 173.5 \\
        SLAHMR~\cite{ye2023decoupling} & 79.1 & - & - & - & - & 25.8 \\
    \midrule
        Hand4Whole~\cite{moon2022accurate} & 71.0 & 59.8 & 127.6 & 48.0 & 41.2  & 27.2 \\
        OSX-L~\cite{lin2023one} & 66.5 & 54.6 & 115.7 & 50.5 & 41.0  & 24.7 \\
        SMPLer-X-B~\cite{cai2023smpler} & 47.1 & 40.7 & 72.7 & 43.7 & 32.4  & 18.9 \\
        \name & \textbf{46.9} & \textbf{39.0} & \textbf{64.7} & \textbf{41.0} & \textbf{26.3} & \textbf{11.6} \\
  \bottomrule
  \end{tabular}
\end{table}

\setlength{\tabcolsep}{4pt}

\begin{table}[t]
  \caption{More camera-frame evaluations on \textbf{EMDB1} and \textbf{3DPW}. Compared to existing mainstream EHPS methods, \name recovers meaningful human depths (T-PVE) and achieves lower acceleration errors (Accl.).}
  \label{tab:cam_space_emdb1_3dpw}
  \centering
  \vspace{-2mm}
  \begin{tabular}{@{}lrrrrrrrr@{}}
    \toprule
        \multirow{2}{*}{Method} & \multicolumn{4}{c}{EMDB1~\cite{kaufmann2023emdb}} & \multicolumn{4}{c}{3DPW~\cite{von2018recovering}} \\
        \cmidrule(lr){2-5} \cmidrule(lr){6-9}
         & PA-PVE$\downarrow$ & PVE$\downarrow$ & T-PVE$\downarrow$ & Accl.$\downarrow$ 
         & PA-PVE$\downarrow$ & PVE$\downarrow$ & T-PVE$\downarrow$ & Accl.$\downarrow$ \\
    \midrule
        Hand4Whole~\cite{moon2022accurate}        & 99.5 & 143.1 & 36851.8 & 34.2 & 81.7 & 124.7 & 30279.0 & 31.0 \\
        OSX-L~\cite{lin2023one}      & 93.3 & 134.0 & 45526.0 & 30.3 & 76.9 & 117.8 & 38472.2 & 24.9 \\
        SMPLer-X-B~\cite{cai2023smpler} & 68.2 & 99.3 & 41298.0 & 24.4 & \textbf{62.6} & 95.6 & 32532.0 & 24.8 \\
        \name      & \textbf{61.0} & \textbf{91.2} & \textbf{140.2} & \textbf{18.4} & 62.8 & \textbf{91.9} & \textbf{260.8} & \textbf{20.3} \\
  \bottomrule
  \end{tabular}
\end{table}

\setlength{\tabcolsep}{4pt}

\begin{table}[tb]
    \vspace{-3mm}
  \caption{Results of camera-frame methods on \textbf{\dataset}. \name is on par with SMPLer-X but produces a lower acceleration error.}
  \label{tab:cam_space_whac_a_mole}
  \centering
  \vspace{-3mm}
  \begin{tabular}{@{}lrrrrrrr@{}}
    \toprule
          & PA-MPJPE$\downarrow$ & PA-PVE-all$\downarrow$ & PVE-all$\downarrow$ & PVE-hand$\downarrow$ & PVE-face$\downarrow$  & Accl.$\downarrow$ \\
    \midrule
        OSX-L~\cite{lin2023one} & 90.1 & 88.1 & 155.7 & 83.3 & 85.0  & 38.9 \\
        SMPLer-X-B~\cite{cai2023smpler} & 76.7 & 74.8 & \textbf{116.2} & \textbf{70.6} & \textbf{63.1}  & 44.0 \\
        \name & \textbf{76.5} & \textbf{74.8} & 117.8 & 77.7 & 63.2 & \textbf{31.2} \\
  \bottomrule
  \end{tabular}
\end{table}

In \cref{tab:cam_space_egobody}, it is shown that \name outperforms existing SoTAs. We highlight that 1) \name archives immense T-PVE-all improvement, which captures absolute depth estimation from humans to cameras. This is because \name formulates the subject distance to the camera. 2) With temporal information embedded in the EHPS module, \name attains substantial reductions in acceleration error (Accl.) compared to previous single-frame SoTAs. Moreover, temporal cues also lead to significant performance gains in hand and face estimation. In \cref{tab:cam_space_emdb1_3dpw}, we further evaluate \name on EMDB and 3DPW, where the plausibility of camera-frame human translation estimation and the significance of temporal modeling are validated again.

Albeit \dataset is mainly designed for world-grounded evaluation of human and camera pose sequences, we evaluate \name's performance under the camera-frame setting on \dataset in \cref{tab:cam_space_whac_a_mole}. Similar to previous experiments, it is observed that \name is on par with SMPLer-X with better performance on the acceleration error.

\subsection{Ablation Study}
\setlength{\tabcolsep}{4pt}

\begin{table}[t]
\centering
\begin{minipage}{0.49\linewidth}
    \caption{\textbf{Ablation} on key components. DPVO represents visual odometry, MV represents \mv.}
      \label{tab:ablation}
      \centering
      \vspace{-2mm}
      \resizebox{\textwidth}{!}{
      \begin{tabular}{@{}lrrrr@{}}
        \toprule
         Method & WA-MPJPE$\downarrow$ &	H-ATE$\downarrow$ & C-ATE$\downarrow$ & C-AS \\
        \midrule
        DPVO & 376.0 & 177.8 & \textbf{14.8} & 5.10 \\
        MV & 233.2 & 129.9 & 134.1 & \textbf{1.10} \\
        MV + DPVO & \textbf{142.2} & \textbf{76.7} & \textbf{14.8} & 1.40 \\
      \bottomrule
      \end{tabular}
      }
\end{minipage}
\hfill
\begin{minipage}{0.49\linewidth}
  \caption{\textbf{Ablation} on intrinsic sources. A reasonable intrinsic drastically improve human root translation estiamtion.}
  \label{tab:ablation_intrinsics}
  \centering
  \vspace{-2mm}
  \resizebox{\textwidth}{!}{
  \begin{tabular}{@{}lrrr@{}}
    \toprule
     & T-MPJPE$\downarrow$ & W-MPJPE$\downarrow$ & WA-MPJPE$\downarrow$ \\
    \midrule
    Dummy(5,000) & 36020.4 & 6239.9 & 604.6 \\
    Assumed~\cite{kissos2020beyond} & 179.7 & 391.2 & 144.0  \\
    GT & \textbf{100.3} & \textbf{389.4} & \textbf{142.2}  \\
  \bottomrule
  \end{tabular}
  }
\end{minipage}
\end{table}

We evaluate the necessity of the key components in ~\cref{tab:ablation}. It is observed that using visual odometry alone (body trajectory depends on estimated camera trajectory) leads to accurate camera trajectory shape (lowest camera trajectory error) but lacks accurate scale (alignment scale is far from 1.0). Using MotionVelocimeter alone (camera trajectory depends on estimated body trajectory), however, results in very accurate scale recovery and better body trajectory error. \name leverages the scale recovery ability of MotionVelocimeter and visual odometer, achieving high-quality body and camera trajectories with only a slight decline in scale accuracy.

\subsection{Visualization}
\begin{figure}[tb]
  \centering
  \vspace{-3mm}
  \includegraphics[width=\textwidth]{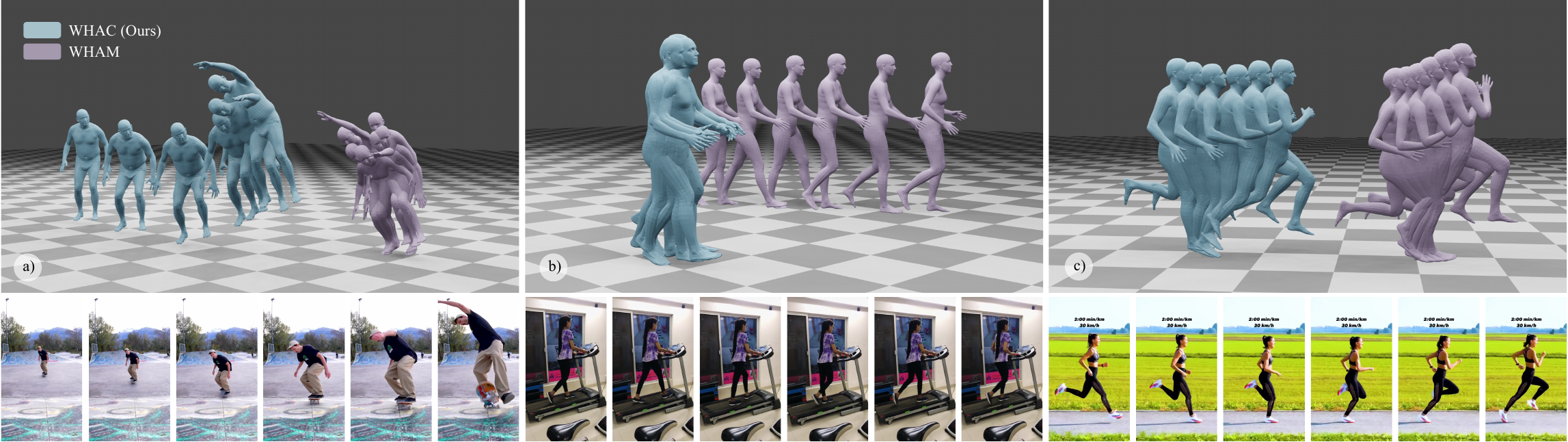}
  \vspace{-4mm}
  \caption{\textbf{Visualization} on in-the-wild hard cases. \name leverages human-camera-scene collaboration to resolve cases where motion prior alone would fail: a) Skateboarding and b) Treadmill. c) \name can also handle fast cases.}
  \label{fig:itw_input_compare}
\end{figure}

We highlight that \name is the first regression-based, whole-body method that simultaneously predicts camera and human trajectories. We highlight that the camera provides supplementary cues to human motions. In \cref{fig:itw_input_compare} a) and b), we test two corner cases where the human motion itself can be misleading: when human pose appears stationary but there is root movement in the world coordinates (\eg, skateboarding), and when human pose clearly indicates motion but there is no root movement in the world coordinates (\eg, running on a treadmill). Our formulation considers both motion and camera cues to predict the correct trajectories whereas WHAM fails, which leverages foot contact and locking but no camera information in human's global trajectory estimation. We also show complicated scenarios on c) in-the-wild video from TikTok, which features a fast-moving object. Our \mv can estimate reasonable root movement, whereas WHAM's contact estimation and foot-locking results in a floating subject. More visualizations are included in the \supp.

\section{Conclusion}
\label{sec:conclusion}
In conclusion, we present \name, the pioneering regression-based EHPS method that jointly recovers human motions and camera trajectories in the world coordinate system. Moreover, our \dataset serves as a useful benchmark for the evaluation of world-grounded EHPS methods. \name achieves SoTA performance on both standard benchmarks and our proposed \dataset, demonstrating strong potentials for downstream applications.

\noindent \textbf{Limitations.} \dataset includes a rich collection of multiperson scenarios that may require special algorithm designs to tackle close interaction and occlusions, which \name lacks. We leave this to the future work.

\noindent \textbf{Potential negative societal impact.} \name may be used for unwarranted surveillance as it recovers human trajectories in the world frame.

\clearpage
\bibliographystyle{plain}
\bibliography{neurips_2023}

\clearpage
\appendix
\section{Overview}
\label{sec:overview}

Given the space constraints in the main paper, we provide additional information and details in this supplementary material: more results visualization on the EMDB dataset and \dataset dataset in \Sec \ref{sec:visualization}; further elaboration on the \mv in \Sec \ref{sec:motion_velocimeter}; the average inference speed in \Sec \ref{sec:inference_speed}; detailed training procedures in \Sec \ref{sec:trainig_details}, explanations of the adaptations applied to camera space EHPS methods during evaluations in \Sec \ref{sec:adaptations}, and the step-by-step comprehensive formulations for trajectory transformations in \Sec \ref{sec:formulation}.
\section{Results Visualization}
\label{sec:visualization}
\begin{figure}[!b]
  \centering
  \includegraphics[width=\textwidth]{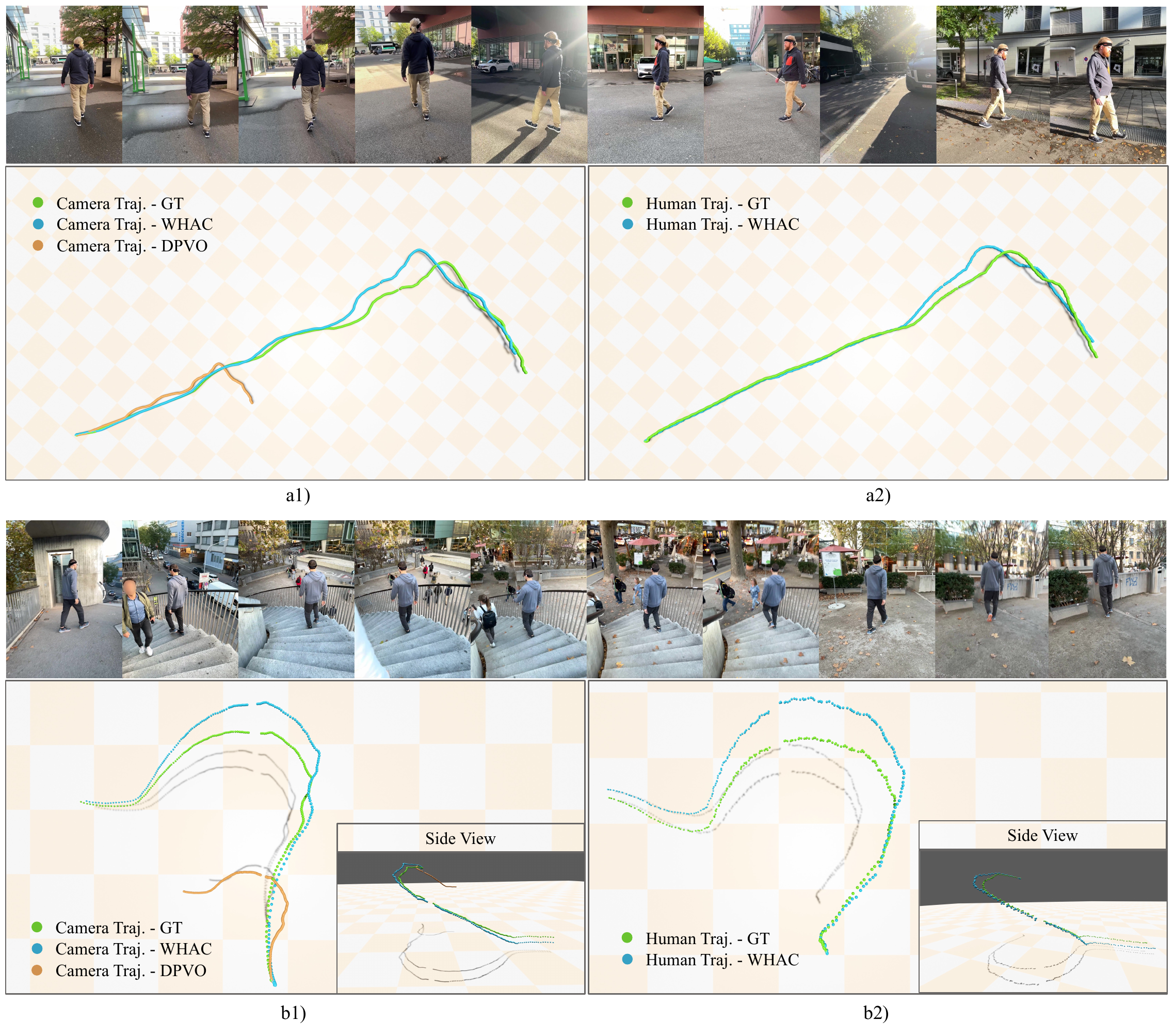}
  \vspace{-4mm}
  \caption{\textbf{Visualization} of world space results on the EMDB dataset. a1) and b1) depict camera trajectories, while a2) and b2) illustrate human trajectories. Notably, in sequence b, the human is descending stairs, and WHAC effectively captures the global trajectory, indicating a downward direction besides recovering the absolute trajectory scale in the world space. The grid size in the plots is 2m.}
  \label{fig:vis_emdb}
\end{figure}
\begin{figure}[t]
  \centering
  \includegraphics[width=\textwidth]{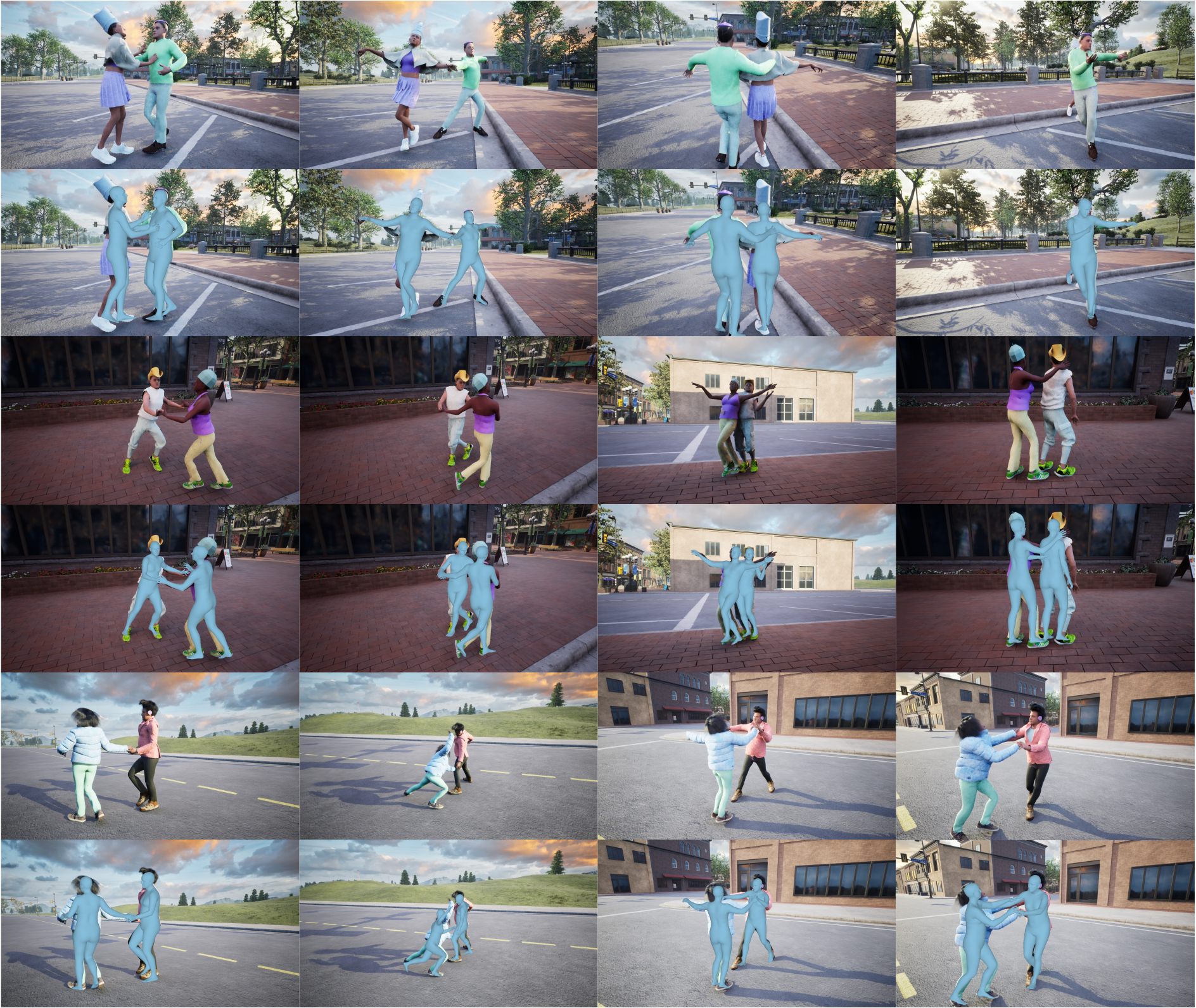}
  \vspace{-4mm}
  \caption{\textbf{Visualization} of camera space results on \dataset dataset. Each sample comprises two rows: the first row displays the original input frames from the sequence, while the second row overlays the SMPL-X results. This visualization showcases WHAC's performance on challenging scenes, including sequences with severe occlusions, intricate human interactions, and dynamic dancing poses.}
  \label{fig:vis_whac_a_mole}
\end{figure}

In Fig. \ref{fig:vis_emdb}, we visualize the camera and human trajectories of two sequences from EMDB2 dataset. The sequences consist of 2.7k frames for and 1.1k frames respectively. Notably, WHAC demonstrates its capability to accurately recover trajectory and scale in the world space, even for lengthy sequences. Moreover, in Fig. \ref{fig:vis_emdb}b1) and Fig. \ref{fig:vis_emdb}b2), WHAC effectively captures the downward trajectory as the depicted human descends stairs.

Besides human and camera trajectory estimation in the world space, we present the visualization of the camera space results in Fig. \ref{fig:vis_whac_a_mole}. This visualization includes various scenarios such as severe occlusions, close interactions, body contact between subjects, and challenging dancing poses, which serve as representative cases for the \dataset dataset. Even without specific training or finetuning on \dataset dataset, WHAC demonstrates strong abilities in handling pose estimation and depth recovery in camera space for various scenarios. However, challenges persist in the recovery of multi-human interactions and contact between body parts in the world space.
\section{\mv}
\label{sec:motion_velocimeter}

\begin{figure}[t]
  \centering
  \includegraphics[width=\textwidth]{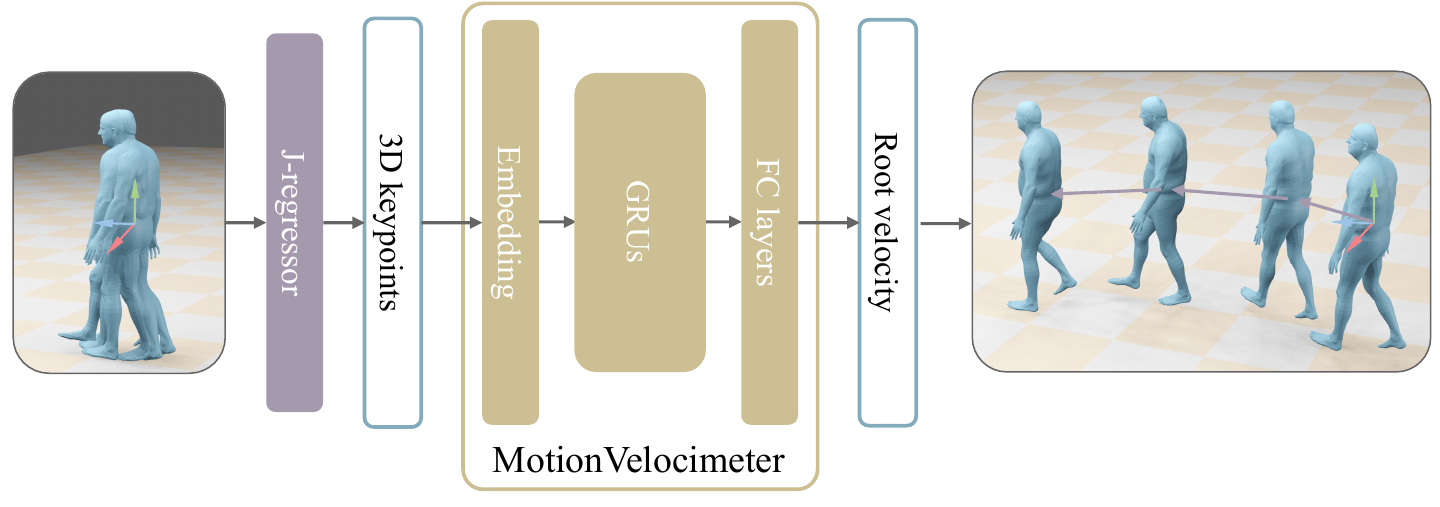}
  \caption{\textbf{Illustration of \mv module}. The inputs are canonicalized 3D joints regressed from SMPL-X meshes, and the outputs are root velocities in the canonical space.}
  \label{fig:motion_velocimeter}
  \vspace{-4mm}
\end{figure}

We present the architecture of \mv in Fig. \ref{fig:motion_velocimeter}. The inputs to \mv consist of canonicalized 3D joints, which are derived from SMPL-X meshes and positioned within the canonical space of the sequence's initial frame. The model's outputs are root velocities corresponding to the previous frame within the canonical space. In contrast to the motion encoder and trajectory decoder that takes 2D keypoints as input in WHAM\cite{shin2023wham}, 3D joints retain spatial information that is critical to velocity estimation in the world-frame for absolute scale recovery. Utilizing 3D joints enhances the model's ability to capture and interpret complex movement patterns, offering a more comprehensive representation of spatial and temporal dynamics within the world-grounded environment.

\section{Inference speed}
\label{sec:inference_speed}
\setlength{\tabcolsep}{4pt}

\begin{table}[t]
  \caption{Inference speed of core modules in frame per second (FPS). *Denotes the inference speed without the replaceable, off-the-shelf modules.}
  \label{tab:inference_speed}
  \centering
  \begin{tabular}{@{}lcccccc@{}}
    \toprule
     Method & GLAMR\cite{yuan2022glamr} &SLAHMR\cite{ye2023decoupling} & PACE\cite{kocabas2023pace} & WHAM\cite{shin2023wham} & WHAC & WHAC*\\
    \midrule
    FPS & 2.4 & 0.04 &  2.1  & 200 & 165 & 2500 \\

  \bottomrule
  \end{tabular}
\end{table}

As indicated in Table \ref{tab:inference_speed}, we assess the inference speed of both the core modules (excluding real-time human detection and visual odometry) following the protocol applied in WHAM\cite{shin2023wham}, as well as the inference speed without off-the-shelf modules, which only includes \mv and the scale recovery of human and camera trajectories. As a regression-based method, the inference speed with WHAC has notably faster inference speeds compared to optimization-based methods.  Its efficiency enables it to meet the real-time speed requirement.
\section{Training Details}
\label{sec:trainig_details}

To enhance temporal consistency and smoothness in camera frame results on temporal datasets, we finetune SMPLer-X-B\cite{cai2023smpler}. This involves incorporating GRUs\cite{cho2014learning} between the ViT-B backbone and the regression heads of the SMPLer-X-B model. We finetune the model on 4 $\times$ V100 GPUs using EgoBody\cite{zhang2022egobody}, 3DPW\cite{von2018recovering}, and EMDB\cite{kaufmann2023emdb} datasets. We set the minimum learning rate to $1 \times 10^{-6}$ for 10 epochs, with a sequence length of 32 frames per batch.

For training \mv, we freeze the finetuned SMPLer-X-B model and initialize the learning rate to $1 \times 10^{-3}$. We employ a step decay learning rate scheduler, reducing the learning rate by $\gamma=0.1$ every 2 epochs. The \mv is trained on 4 $\times$ V100 GPUs over 8 epochs.
\section{Adaptations for camera frame EHPS methods}
\label{sec:adaptations}

In Table 2 and Table 3 of the main paper, we employ world-grounded adaptations to camera frame methods such as OSX\cite{lin2023one} and SMPLer-X\cite{cai2023smpler} using the approach described in Sec. 3.2. This method enables the recovery of camera-space root depth, ensuring a fair comparison with other world-grounded methods. Without the adaptations, camera frame methods face limitations when compared to world-grounded methods, due to the high T-MPJPE observed in the camera frame.

\section{Comprehensive Formulations}
\label{sec:formulation}

Given the space constraints, we present only the finalized and general formulations in Sec. 3.3 and Sec. 3.4 of the main paper. Here, we include the comprehensive step-by-step formulations of the trajectory transformations for clarity.

\subsection{Canonicalization}
We briefly explain the transformation for canonicalization in Eq. 6 in the main paper. Here we provide the comprehensive formulation:
\begin{equation}
    T^{cano} =T^{cano}_{w, i} = [R^{cano} | t^{cano}] = [R^{cano}_{w,i} | t^{cano}_{w,i}],
\end{equation}
where $T^{cano}$ is a generalized symbol for the transformation from world coordinate system to canonical coordinate system, $i$ denotes the $i^{th}$ frame in the sequence.

For the first ($0^{th}$) frame in the sequence:
\begin{equation}
    R^{cano}_{w,0} = (R^{w}_{c,0} \times \theta^{c}_{go,0})^{-1} = (\theta^{c}_{go,0})^{-1},
    t^{cano}_{w,0} = - p^{w}_{0},
\end{equation}
where $R^{w}_{c,0}$ is $0^{th}$ camera rotation in the world frame estimated from visual odometry, $\theta^{c}_{go,0}$ is $0^{th}$ global orientation estimated in the camera space, $- p^{w}_{0}$ is the pelvis joint of $J^{w}_{0}$ for the first frame.

For every frame in the sequence:
\begin{equation}
    R^{cano}_{w,i} = (R^{w}_{c,0} \times \theta^{c}_{go,0})^{-1} = (\theta^{c}_{go,0})^{-1},
    t^{cano}_{w,i} = - p^{w}_{i}.
\end{equation}
We use the pelvis translation $- p^{w}_{i}$ for the corresponding frame while using the camera rotation and global orientation of the first frame for the entire sequence in canonicalization. This is to retain the human rotation in the world coordinate system between frames for reliable trajectory estimation.

\subsection{Derive Camera Trajectories from Human Trajectories}

In Sec. 3.4 in the main paper, we briefly explain the process of deriving camera trajectories $T^{w}_{c,derived}$ from human trajectories $T^{w}_{h}$ in Eq. 9. We append the full formulation for this process:

\begin{equation}
    T^{w}_{c,derived} =(T^{cano})^{-1} \times  T^{cano}_{h} \times (T^{c}_{h})^{-1} = T^{w}_{cano} \times T^{cano}_{h} \times T^{h}_{c} = T^{w}_{h} \times T^{h}_{c},
\end{equation}

\begin{equation}
    T^{w}_{h} = [R^{w}_{h} | t^{w}_{h}],
\end{equation}

\begin{equation}
    T^{h}_{c} = [R^{h}_{c} | t^{h}_{c}] = [\theta^{c}_{go,i}|t^{c}_{h,i}]^{-1}, t^{h}_{c} = -(\theta^{c}_{go,i})^{-1} \times p^{c}_{i},
\end{equation}
where $T^{w}_{h}$ is the output human trajectories in the world coordinate system from \mv,  $T^{h}_{c}$ is the inverted human root transformation in the camera coordinate system, $\theta^{c}_{go,i}$ and $p^{c}_{i}$ is the $i^{th}$ global orientation and pelvis in camera space respectively. 

To further process the camera rotations $R^{w}_{c,derived}$ and camera moving trajectories $t^{w}_{c,derived}$ separately, we re-write the transformations mentioned above in the form of $T = [R | t]$:

\begin{equation}
    T^{w}_{c,derived} = [R^{w}_{c,derived} |t^{w}_{c,derived}]
    = [R^{w}_{h} | t^{w}_{h}] \times [R^{h}_{c} | t^{h}_{c}]
    = [R^{w}_{h}R^{h}_{c} | R^{w}_{h}t^{h}_{c} + t^{w}_{h}],
\end{equation}

\begin{equation}
    R^{w}_{c,derived} = R^{w}_{h} \times R^{h}_{c},
    t^{w}_{c,derived} = R^{w}_{h} \times t^{h}_{c} + t^{w}_{h}, 
\end{equation}
where $t^{w}_{c,derived}$ is the human-derived camera trajectory in the world coordinate system with absolute scale. The scale recovery is explained in Eq. 11 in the main paper. 
 
\subsection{Derive Human Trajectories from Camera Trajectories}
In Sec. 3.4 and Eq. 12, we explain the process of deriving the human trajectories $T^{w}_{h,final}$ from scale-recovered camera trajectories $T^{w}_{c,final}$:
\begin{equation}
     T^{w}_{c,final}=[R^{w}_{c} | t^{w}_{c,final}],
\end{equation}

\begin{equation}
    T^{w}_{h,final} = [R^{w}_{h,final} | t^{w}_{h,final}] = T^{w}_{c,final} \times T^{c}_{h},
\end{equation}
where $T^{w}_{c,final}$ is the scale-recovered VO-estimated camera trajectories. We empirically find that $R^{w}_{c}$ estimated with visual odometry is accurate and can be used in the final camera trajectory $T^{w}_{c,final}$. The scale recovery process via Umeyama alignment\cite{umeyama1991least} for $t^{w}_{c,final}$ is explained in Eq. 11 in the main paper. $T^{c}_{h}$ is the human root transformation in the camera coordinate system. 

By first deriving camera trajectories from \mv-estimated human trajectories, and followed by deriving human trajectories from scaled VO-estimated camera trajectories, we obtain human trajectory $T^{w}_{h,final}$ and camera trajectory $T^{w}_{c,final}$, both in the world coordinate system and with absolute scales. 

\end{document}